\documentclass{article}

\usepackage[preprint]{neurips_2025}

\usepackage[utf8]{inputenc} 
\usepackage[T1]{fontenc}    
\usepackage{hyperref}       
\usepackage{url}            
\usepackage{booktabs}       
\usepackage{amsfonts}       
\usepackage{nicefrac}       
\usepackage{microtype}      
\usepackage{xcolor}         
\usepackage{amsmath}
\usepackage{graphicx}
\usepackage[capitalize,noabbrev]{cleveref}
\usepackage{svg}

\title{EquiHGNN: Scalable Rotationally Equivariant Hypergraph Neural Networks}

\author{%
  Tien Dang\thanks{Work completed during an internship at UAB} \\
  Department of Computer Science \\
  The University of Alabama at Birmingham \\
  Birmingham, Alabama 35294, USA \\
  \texttt{tien.danganh02@gmail.com} \\
  \And
  Truong-Son Hy\thanks{Corresponding Author} \\
  Department of Computer Science \\
  The University of Alabama at Birmingham \\
  Birmingham, Alabama 35294, USA \\
  \texttt{thy@uab.edu}
}

\begin{document}

\maketitle
\begin{abstract}

Molecular interactions often involve high-order relationships that cannot be fully captured by traditional graph-based models limited to pairwise connections. Hypergraphs naturally extend graphs by enabling multi-way interactions, making them well-suited for modeling complex molecular systems. In this work, we introduce \textbf{EquiHGNN}, an \textbf{Equi}variant \textbf{H}yper\textbf{G}raph \textbf{N}eural \textbf{N}etwork framework that integrates symmetry-aware representations to improve molecular modeling. By enforcing the equivariance under relevant transformation groups, our approach preserves geometric and topological properties, leading to more robust and physically meaningful representations. We examine a range of equivariant architectures and demonstrate that integrating symmetry constraints leads to notable performance gains on large-scale molecular datasets. Experiments on both small and large molecules show that high-order interactions offer limited benefits for small molecules but consistently outperform 2D graphs on larger ones. Adding geometric features to these high-order structures further improves the performance, emphasizing the value of spatial information in molecular learning. Our source code is available at \url{https://github.com/HySonLab/EquiHGNN/}.

\end{abstract}

\section{Introduction}\label{sec:introduction}
Molecular systems exhibit complex, high-order interactions, including conjugated \(\pi\)-systems, hydrogen bonding networks, and ring strain effects~\cite{david2020molecular, wigh2022review}. Taking advantage of the inherent benefits of graph-based representations, Graph Neural Networks (GNNs) have been widely used to model molecular interactions~\cite{gilmer2017neural} due to their ability to efficiently learn relational structures. GNNs leverage message passing mechanisms to aggregate information from neighboring atoms, making them well suited to encode local connectivity and bond-based interactions~\cite{zitnik2018modeling, gligorijevic2021structure}. However, standard GNNs primarily model pairwise relationships between nodes, limiting their ability to represent the multi-body dependencies inherent in molecular interactions. Furthermore, they often lack explicit geometric information, such as spatial coordinates, bond angles, and torsional relationships, that is crucial for accurately capturing the three-dimensional structure and properties of molecules.

\begin{figure}[ht]
    \centering
    \label{fig:conj_bipartite}
    \includegraphics[width=0.85\textwidth]{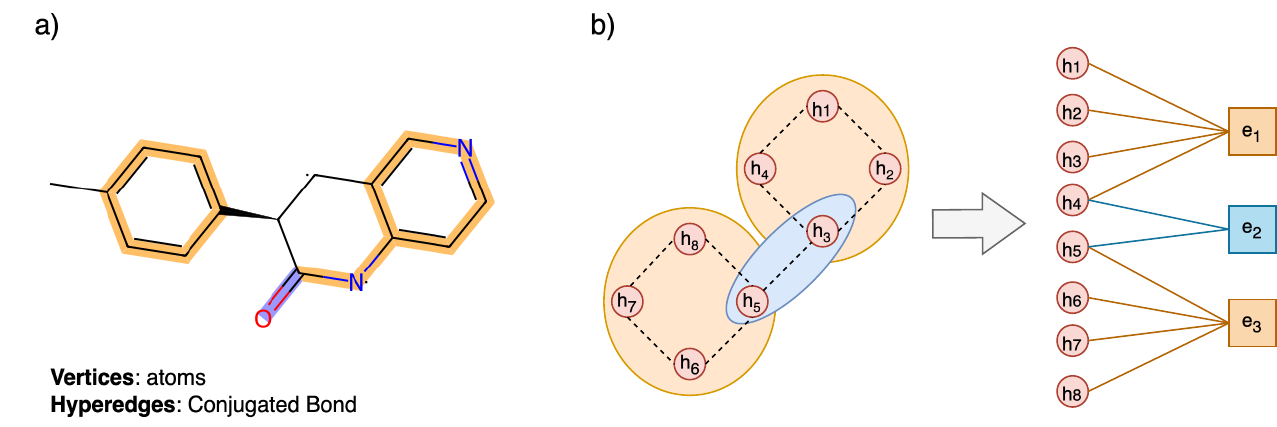}
    \caption{\textit{a)} Illustration of a hypergraph constructed from a molecule, where vertices represent atoms and hyperedges represent conjugated bonds, highlighted in blue and orange. \textit{b)} Hypergraph to Bipartite representations.}
\end{figure}

Topological Deep Learning (TDL)~\cite{hajij2022topological, pmlr-v235-papamarkou24a, papillon2023architectures} offers a robust framework to overcome the limitations of GNN by integrating higher-order structures that extend beyond simple node-to-node connections. Techniques such as simplicial complexes~\cite{bodnar2021weisfeiler}, cell complexes~\cite{CWbodnar2021weisfeiler}, combinatorial complexes~\cite{hajij2022topological, hajij2022higher} and hypergraphs~\cite{HNHN2020, chien2021you, chen2024molecular} enrich traditional graph representations by capturing complex multi-body interactions intrinsic to molecular systems. By embedding these topological structures within deep learning architectures, we can achieve a more holistic and principled understanding of molecular interactions. This approach has already led to state-of-the-art performance in a variety of machine learning tasks~\cite{HNHN2020, barbarossa2020topological, hajij2022higher, chen2022bscnets}, and holds significant promise to advance research and applications in the applied sciences and beyond.

In addition to topological considerations, molecular interactions are inherently governed by geometric constraints. Geometric Deep Learning (GDL)~\cite{bronstein2021geometric} incorporates geometric priors, such as structural and symmetry information about the input space, which are essential to maintain spatial symmetries and to ensure that the learned representations accurately reflect the physical properties of molecules. A key requirement in molecular modeling is the equivariance to transformations such as rotations, translations, and reflections. Recent advancements in equivariant architectures enable the integration of geometric constraints into deep learning models, ensuring that molecular representations align with the principles of three-dimensional spatial organization~\cite{schutt2017schnet, NEURIPS2019_03573b32, batzner20223,zhang2023universal, wang2024enhancing, equiformer_v2, Huang2024ProteinNucleicAC}.

Although previous work has explored integrating symmetry into simplicial~\cite{eijkelboom2023n} and combinatorial complexes\cite{battiloro2024n},  incorporating equivariant features into hypergraphs offers a more expressive and adaptable framework for modeling complex multi-body interactions in molecular systems. Unlike other complex-based representations, hypergraphs provide a more direct and scalable approach to capturing these interactions, enabling more intuitive molecular modeling. Given these advantages, in this study we focus on hypergraphs to model high-level interactions in molecular systems.

We introduce EquiHGNN, a unified framework that seamlessly integrates topological and geometric learning to model molecular interactions while preserving geometric consistency. Instead of designing a complex message-passing framework, we initialize the hypergraph features with symmetry-aware geometric representations and invariant scalar information, enabling the model to effectively capture structural and functional properties without sacrificing equivariance. Compared to baseline models that do not incorporate symmetry into hypergraph representations, our approach encodes both scalar and geometric features within a unified framework. 

We evaluated our model on QM9~\cite{ramakrishnan2014quantum}, OPV~\cite{st2019message}, PCQM4Mv2~\cite{hu2021ogb}, and Molecule3D~\cite{xu2021molecule3d} to assess both accuracy and scalability. While QM9 and OPV contain small molecules, PCQM4Mv2 and Molecule3D test performance on large-scale graphs. Our model achieved competitive results across all datasets, demonstrating strong generalization and scalability to complex molecular structures.

Our contributions are as follows. 
\begin{itemize}
    \item We propose Equivariant Hypergraph Neural Network (EquiHGNN) for molecular property prediction that effectively captures both scalar and geometric features through hypergraph representations. Its modular and intuitive design enables seamless integration with existing frameworks and allows reuse of publicly available models.
    
    \item We conducted an empirical study to analyze the impact of various architectural choices and configurations, exploring different equivariant approaches. Specifically, we examine the use of EGNN~\cite{satorras2021n} in the Euclidean domain, Equiformer~\cite{liao2023equiformer} in the Fourier domain, and FAFormer~\cite{Huang2024ProteinNucleicAC} in the frame domain. 
    
    \item We observe that high-order interactions in hypergraphs, although slightly less effective on small molecules, consistently enhance performance on large-scale graphs. Additionally, integrating 3D geometric information into the hypergraph significantly improves the accuracy of the model.
\end{itemize}

\section{Related Work}

\subsection{Graph Neural Network}

Graph Neural Networks (GNNs) have been widely developed to improve representation learning in graph-structured data, enabling effective modeling of relational and structural information \cite{zhou2020graph}. GraphSAGE~\cite{hamilton2017inductive} introduced an inductive framework that aggregates information from the local neighborhood of a node, allowing generalization to unseen graphs. More general, Message Passing Neural Networks (MPNNs)~\cite{gilmer2017neural} are a foundational class of GNNs that iteratively update node representations by aggregating and transforming information from their neighbors, enabling effective learning on graph-structured data. Graph Convolutional Networks (GCN)~\cite{kipf2016semi} leveraged spectral graph theory to perform efficient message passing through neighborhood-based feature propagation. Graph Isomorphism Networks (GIN)~\cite{xu2018powerful} maximized expressive power by using sum aggregation, making them as discriminative as the Weisfeiler-Lehman test for graph isomorphism. Graph Attention Networks (GAT)~\cite{GAT2018} integrated attention mechanisms to dynamically weight neighbor contributions, improving the model’s ability to capture important structural dependencies. However, these models operate on discrete graph structures, lack inherent equivariance or invariance to geometric transformations such as rotations, translations, and reflections, and can only model pairwise interactions, making them impractical for capturing high-order molecular interactions such as conjugate bonds. This work addresses these limitations by proposing EquiHGNN, a framework that incorporates both geometric awareness and higher-order interaction modeling.

\subsection{Geometric Graph Neural Networks}

Geometric graph \cite{bronstein2021geometric} is a special kind of graph with geometric information, e.g. the positions of the atoms in 3D coordinates, encapsulating rich directional information that depicts the geometry of the system, making the system ineffectively processed by GNNs. Researchers propped a variety of Geometric Graph Neural Networks quipped with invariant/equivriant properties to better characterize the geometry of geometric graph.

Many tasks require models to be invariant under Euclidean transformations, which is often achieved by converting equivariant coordinates into invariant scalars. Early works like Cormorant \cite{NEURIPS2019_03573b32} introduced the idea of using covariant tensorial representations for molecular graphs, ensuring that the learned features transform predictably under rotations and translations. Using spherical harmonics and tensor contractions, Cormorant demonstrated how symmetry-preserving architectures can substantially improve molecular property predictions.  SchNet~\cite{schutt2017schnet} uses continuous filter convolutions with filter weights conditioned on relative distances but lacks directional encoding. DimeNet~\cite{gasteiger_dimenet_2020} addresses this by introducing directional message passing, incorporating both distances and angles between adjacent edges. GemNet~\cite{gasteiger2021gemnet} extends this further by incorporating dihedral angles, enabling more expressive two-hop directional message passing based on quadruplets of nodes.

Equivariant graph neural networks, on the contrary, simultaneously update invariant and equivariant features, as many tasks require equivariant output \cite{GraphRepresentationLearning:4761}. EGNN~\cite{satorras2021n}, a well-known scalarization-based model, constrains messages to invariant distances and multiplies them by relative coordinates to ensure equivariant updates. Frame Averaging (FA) \cite{puny2021frame, duval2023faenet} ensures equivariance by encoding coordinates in multiple reference frames and averaging their representations. 
Since summing over all group elements is computationally difficult, 
, FA selects a representative subset using a frame function~\cite{puny2021frame}. This method has been further explored in material design, offering a scalable alternative to traditional equivariant architectures~\cite{duval2023faenet}. FAFormer~\cite{Huang2024ProteinNucleicAC} incorporates the Transformer with frame averaging within each layer, offers superior performance in the prediction of contact maps and the detection of aptamers.

Spherical harmonics-based models use functions derived from spherical harmonics and irreducible representations, leveraging tensor product operations to ensure equivariant data transformations \cite{s.2018spherical, NEURIPS2019_03573b32}. Tensor Field Network (TFN)~\cite{thomas2018tensor} and NequIP\cite{batzner20223} utilize equivariant graph convolutions with linear messages derived from tensor products, with NequIP further enhancing this approach using equivariant gate activations. The SE(3)-Transformer~\cite{fuchs2020se} extends SEGNN~\cite{brandstetter2022geometric} by replacing equivariant gate activations with equivariant dot product attention for dynamic interaction weighting, while Equiformer~\cite{liao2023equiformer} further enhances it with MLP-based attention, equivariant layer normalization, and regularizations such as dropout and stochastic depth.

As shown in~\cite{joshi2023expressive}, rotationally equivariant GNNs are more expressive than invariant GNNs, especially for sparse geometric graphs. In this work, we focus on equivariant methods, specifically EGNN~\cite{satorras2021n} for the scalarization-based approach, FAFormer~\cite{Huang2024ProteinNucleicAC} for the frame-averaging-based approach, and Equiformer~\cite{liao2023equiformer} for the spherical harmonic-based approach.

\subsection{Topological Deep Learning}

Topological Deep Learning (TDL) \cite{hajij2022topological, pmlr-v235-papamarkou24a} extends beyond traditional graphs by leveraging higher-order structures, enabling a more expressive framework for modeling complex interactions among multiple entities simultaneously. Beyond rotational symmetry, molecular graphs also exhibit rich permutation symmetries in their relational structure. Predicting molecular properties with Covariant Compositional Networks (CCNs) \cite{10.1063/1.5024797} proposed a framework that preserves higher-order permutation equivariance during message passing by modeling interactions as higher-order tensors. This approach enables the network to learn more expressive and physically meaningful representations compared to first-order (pairwise) GNNs, especially when modeling complex molecular systems with many-body interactions. The Weisfeiler-Lehman graph isomorphism test has been extended to simplicial and regular cell complexes ~\cite{bodnar2021weisfeiler,CWbodnar2021weisfeiler}, providing a theoretical foundation for higher-order graph structures. HGNN~\cite{feng2019hypergraph} introduced a spectral-based framework that utilizes the Laplacian hypergraph to pass messages across hyperedges. To improve flexibility, AllSet~\cite{chien2021you} proposed a more general approach that models hypergraphs as multi-sets, employing learnable permutation-invariant set functions for adaptive message aggregation. Compared to baselline pretrained GNNs, MHNN~\cite{chen2024molecular} takes advantage of the hypergraph to achieve better performance under limited training data. CCNN~\cite{hajij2022topological} further advances this direction by introducing Combinatorial Complexes, which capture hierarchical order and enable structured dependencies across multiple levels. A comprehensive review of these advances can be found in ~\cite{papillon2023architectures}.

Despite these developments, there has been limited work that incorporates symetry with topological structures. Recent efforts have explored simplicial complexes with equivariant message passing ~\cite{eijkelboom2023n, liu2024clifford}, integrating symmetry-aware mechanisms into higher-order networks~\cite{liu2024clifford}. Additionally, ETNN~\cite{battiloro2024n} extends equivariant message passing to combinatorial complexes, providing a more structured approach to learning equivariant representations in topological deep learning.

Hypergraphs offer a powerful framework for modeling higher-order interactions, particularly in domains such as molecular modeling and complex systems. However, equivariant hypergraph neural networks remain largely unexplored. This work introduces a novel hypergraph-equivariant framework that inherits the scalability of graph-based methods, making it suitable for large molecular systems, while also incorporating equivariant geometric features to enhance expressive power and robustness.

\section{Method}

This section presents our intuitive approach to modeling high-order interactions with symmetry-aware features, emphasizing key components and detailing their seamless integration for optimal performance.

\begin{figure}
    \centering
    \includegraphics[width=0.99\linewidth]{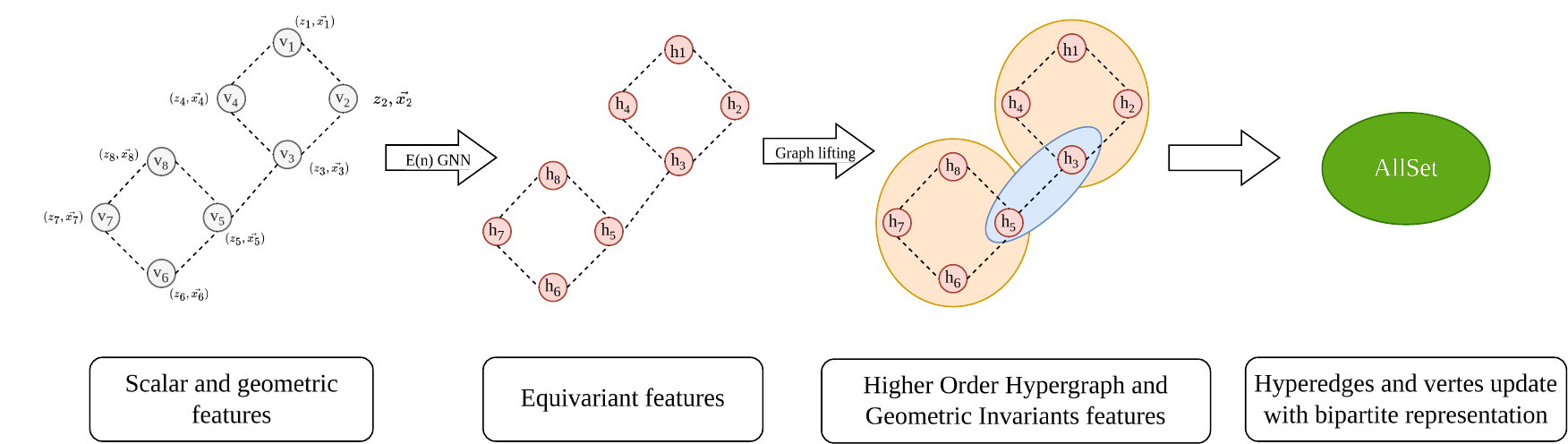}
    \caption{Overview of the Equivariant Hypergraph Neural Network framework.}
    \label{fig:pipeline}
\end{figure}

\subsection{AllSet}


The AllSet framework~\cite{chien2021you} formulates HGNNs using multiset functions, ensuring permutation invariance and expression of the message passing. It models hypergraphs as bipartite graphs, enabling flexible message propagation through two learnable set functions.

Formally, let \(\mathcal{H} = (\mathcal{V}, \mathcal{E})\) be a hypergraph, where \(\mathcal{V}\) is the set of nodes and \(\mathcal{E}\) is the set of hyperedges, each of which connects a subset of nodes. The AllSet framework updates representations through a two-step message-passing mechanism. First, hyperedge embeddings are computed by aggregating features from incident nodes using a set function \( f_{V \to E} \), defined as follows:
\[
Z^{(t+1)}_e = f_{V \to E} \left( V_{e \setminus v, X^{(t)} } ; Z^{(t),v}_{e,:} \right),
\]
where \( X^{(t)} \) represents the features of nodes in iteration \( t \), \( Z^{(t)}_e \) denotes hyperedge embeddings, and \( V_e \) is the set of nodes belonging to hyperedge \( e \). This function aggregates node information into a hyperedge representation while preserving permutation invariance.  

The node features are then updated based on the embeddings of the hyperedges through a second set function \( f_{E \to V} \), which propagates information back from the hyperedges to the nodes:
\[
X^{(t+1)}_v = f_{E \to V} \left( E_{v, Z^{(t+1),v}_e} ; X^{(t)}_{v,:} \right),
\]
where \( E_v \) is the set of hyperedges containing the node \( v \). This formulation allows the message-passing process to flexibly capture complex dependencies between nodes and hyperedges.  

After $T$ steps of the message passing, the hypergraph-level prediciotn is calculated in the readout part on the final hidene states of hyperedges and nodes:
\[
y = \text{MLP} \left( \sum_{v \in G}\textit{X}_{v}^{(T)} \sum_{e \in G}\textit{Z}_{e}^{(T)} \right).
\]
This architecture ensures permutation invariance while allowing expressive transformations of hypergraph features. 

\subsection{Equivariant Hypergraph Neural Network}

Our approach enhances AllSet by initializing node features with both scalar attributes and 3D geometric properties, naturally embedding equivariant information into the model. This simple yet effective design leads to a more expressive and symmetry-aware framework for hypergraph learning.

To explore the best strategy for symmetry-aware representation, we evaluated models across three geometric domains. Scalar-based models such as EGNN~\cite{satorras2021n} preserve pairwise distances and relational structure, ensuring invariance. Frame-based methods such as FAFormer~\cite{Huang2024ProteinNucleicAC} apply frame averaging to enforce equivariance while capturing higher-order geometric patterns. Fourier-based models such as Equiformer~\cite{liao2023equiformer} transform geometric features into the spectral domain to model long-range dependencies. This comparative analysis helps identify the most effective backbone for incorporating geometric symmetry into hypergraph learning.

As illustrated in \cref{fig:pipeline}, we first extract symmetry-aware embeddings using an equivariant model (e.g., EGNN or Equiformer). These embeddings are then used as input features in a hypergraph module that captures complex higher-order interactions. The hyperedges embedding is concatenated with the vertexes features and passed through several MLP layers to generate the final predictions. Although the pipeline remains conceptually simple, it significantly improves performance over other baselines.

\section{Experiment}

Density Functional Theory (DFT) is a widely-used quantum mechanical method for predicting molecular properties, such as structure, reactivity, and responses to electromagnetic fields. Although DFT provides high accuracy, its computational cost increases rapidly with the size of the system, making it impractical for large-scale molecular screening. As a result, DFT calculations can be time-consuming, often requiring several hours for even small molecules. This limitation makes it challenging to explore large chemical spaces or conduct extensive property predictions.

In this study, we utilize the QM9~\cite{ramakrishnan2014quantum} and OPV~\cite{st2019message} datasets for molecular property prediction. QM9 consists of small organic molecules and is widely used to benchmark fundamental molecular properties. The OPV dataset focuses on organic photovoltaic molecules, which typically contain larger conjugated systems relevant for electronic and optoelectronic applications.

To assess the scalability and expressiveness of our model, we evaluate large-scale graph datasets such as PCQM4Mv2~\cite{hu2021ogb} and Molecule3D~\cite{xu2021molecule3d}, both derived from PubChemQC~\cite{nakata2017pubchemqc}. These datasets are designed to predict critical molecular properties for fields such as drug discovery and materials science, featuring complex molecular structures and substantial graph data. While PCQM4Mv2 is focused primarily on predicting the HOMO-LUMO gap from SMILE strings, Molecule3D emphasizes 3D molecular geometry prediction, integrating molecular properties prediction as a secondary task. 


\subsection{Dataset}

\begin{table}[h]
    \centering
    \caption{Overview of the datasets}
    \renewcommand{\arraystretch}{1.2}
    \label{tab:datasets}
    \begin{tabular}{lcccc}
        \toprule[1pt]\midrule[0.3pt]
        Dataset & Graphs & Task type & Task number & Metric \\
        \midrule
        QM9  & 134K & regression & 12 & MAE \\
        OPV  & 91K & regression & 8  & MAE \\
        PCQM4Mv2  & 3.7M & regression & 1  & MAE \\
        Molecule3D  & 3.9M & regression & 6  & MAE \\
        \midrule[0.3pt]\bottomrule[1pt]
    \end{tabular}
\end{table}

\textbf{QM9}~\cite{ramakrishnan2014quantum} dataset is a widely used reference for the prediction of chemical properties. It comprises approximately 134,000 small organic molecules, each containing up to 29 atoms. The data set includes five atomic species including hydrogen, carbon, oxygen, nitrogen, and fluorine, structured as molecular graphs where the atoms are connected by four types of chemical bonds: single, double, triple, and aromatic. In addition, the 3D coordinates of each atom, measured in angstroms, are provided.

\textbf{Organic Photovoltaic (OPV)}~\cite{st2019message} comprises 90,823 unique molecules, providing their SMILES representations, 3D geometries, and optoelectronic properties computed by DFT calculations. It includes four molecular-level tasks for monomers: the highest occupied molecular orbital (\(\varepsilon_{\text{HOMO}}\)), the lowest unoccupied molecular orbital (\(\varepsilon_{\text{LUMO}}\)), the HOMO-LUMO gap (\(\Delta\varepsilon\)), and the spectral overlap (\(I_{\text{overlap}}\)). Furthermore, OPV features four polymer-level tasks: polymer \(\varepsilon_{\text{HOMO}}\), polymer \(\varepsilon_{\text{LUMO}}\), the polymer energy gap (\(\Delta\varepsilon\)), and the optical LUMO (\(O_{\text{LUMO}}\)).

\textbf{PCQM4Mv2}~\cite{hu2021ogb} is a large-scale quantum chemistry dataset consisting of approximately 3.7 million molecular graphs, derived from the PubChemQC project~\cite{nakata2017pubchemqc}. It is designed for predicting the DFT-calculated HOMO-LUMO energy gap from SMILES representations, and additionally provides 3D structures for the training molecules.

\textbf{Molecule3D}~\cite{xu2021molecule3d} is a large-scale benchmark designed to predict 3D molecular geometries from 2D molecular graphs and to assess their effectiveness in downstream prediction of quantum properties. It supports two main tasks: (1) prediction of DFT-optimized 3D atomic coordinates from SMILES strings or molecular graphs, and (2) prediction of quantum properties such as total energy, HOMO/LUMO energies, and the HOMO–LUMO gap using either ground truth or predicted 3D structures. Each sample includes a SMILES string, molecular graph, 3D coordinates, and quantum properties sourced from PubChemQC~\cite{nakata2017pubchemqc}. In this work, we focus specifically on the prediction of the HOMO-LUMO gap.

\Cref{tab:datasets} provides an overview of the experimental dataset. We use RDKit to identify conjugated bonds, which serve as hyperedges, with atoms as vertices, as illustrated in \cref{fig:conj_bipartite}a. For all experiments, the data are split into training, validation and test sets using an 80-10-10 ratio. The model is trained in the training set, the best model is selected based on the performance in the validation set, and the final evaluation is performed in the test set.

\subsection{Training details}


Equivariant models utilize radial distances, where a larger radius enables the capture of high-level features crucial for complex molecules such as polymers and proteins. In such molecules, long-range interactions, such as electrostatic and hydrophobic effects, are key to determining their electronic and structural properties. A study on scaling GNNs~\cite{airas2024scaling} shows that increasing the number of message-passing layers and the cutoff radius helps GNNs incorporate distant atomic interactions, thereby enhancing expressiveness for large proteins.

However, in transformer-based architectures, a larger radius significantly increases computational costs due to the quadratic scaling of the attention mechanism with the number of nodes. Based on these empirical insights, we adopt a consistent configuration with a 5 Angstrom radius cutoff and 16 neighboring nodes for EGNN, FAFormer, and Equiformer, achieving an optimal balance between expressiveness and computational efficiency. \cref{appendix:model_overview} provides a summary of the model architecture configurations.

We train the models for 400 epochs with a batch size of 16, using the Adam optimizer with a fixed learning rate of \(1 \times 10^{-4}\). Training is carried out on 2xRTX 3060 GPUs, enabling parallel processing for efficiency. The models are optimized to minimize the loss of MSE, and the checkpoint with the lowest MAE validation is selected for the final evaluation on the test set. Our implementation is built using PyTorch Geometric~\cite{FeyLenssen2019}.

\subsection{Results}

We perform experiments comparing our approach to the MHNN baseline~\cite{chen2024molecular}, which represents the hypergraph using a bipartite graph structure, similar to AllSet~\cite{chien2021you}. To incorporate symmetry-aware features, we evaluated three different setups: EGNN~\cite{satorras2021n}, FAFormer~\cite{Huang2024ProteinNucleicAC}, and Equiformer~\cite{liao2023equiformer}.

Furthermore, we compare performance with 2D graph models to assess whether incorporating higher-order interactions with symmetry awareness can improve the model performance. All results show significantly better performance compared to 2D graph models, including GIN~\cite{xu2018powerful} and GAT~\cite{veličković2018graph}, highlighting the importance of capturing geometric and topological information.

The tables below report the MAE scores across all datasets, with bold values indicating the best-performing models and underlined values denoting the second-best. For the PCQM4Mv2 and Molecule3D datasets, only EGNN-MHNN is included among the equivariant integration models, as training FaFormer and Equiformer on these large-scale datasets is prohibitively time-consuming. In particular, in PCQM4Mv2, 3D geometric information is available only for the training set; therefore, all experiments are conducted on this subset to ensure a fair comparison between 2D and 3D graph representations.

\subsubsection{QM9 dataset}

\begin{table}[h]
    \centering
    \caption{MAE on the QM9 test set.}
    \renewcommand{\arraystretch}{1.2}
    \label{tab:qm9_res}
    \resizebox{\textwidth}{!}{
    \begin{tabular}{lcccccc}
        \toprule[1pt]\midrule[0.3pt]

        Task & $\mu $ & $\alpha $ &  $\epsilon_{\textrm{HOMO}}$ & $\epsilon_{\textrm{LUMO}}$ & $\Delta \epsilon$ & $\langle R^2 \rangle$ \\
        
        Units ($\downarrow$) & $\textrm{D}$ & ${a_0}^3$ & meV & meV & meV & ${a_0}^2$ \\
        \midrule
        GIN & 
        \textbf{0.2} $\pm$ \textbf{0.003} & 
        4.09 $\pm$ 0.04 & 
        47.67 $\pm$ 0.4 & 
        99.62 $\pm$ 0.9 & 
        147.87 $\pm$ 1.3 & 
        6279.8 $\pm$ 86.91 \\

        GAT & 
        0.65 $\pm$ 0.006 & 
        6.17 $\pm$ 0.07 & 
        51.56 $\pm$ 0.5 & 
        111.24 $\pm$ 0.9 & 
        158.26 $\pm$ 1.1 & 
        8772.5 $\pm$ 97.43 \\

        MHNN & 
        0.67 $\pm$ 0.005 & 
        9.29 $\pm$ 0.1 & 
        55.38 $\pm$ 0.5 & 
        124.23 $\pm$ 1.07 & 
        166.6 $\pm$ 1.3 & 
        9301.44 $\pm$ 138.369 \\
        
        EGNN-MHNN & 
        0.59 $\pm$ 0.005 & 
        \textbf{2.02} $\pm$ \textbf{0.02} & 
        44.82 $\pm$ 0.3 & 
        92.54 $\pm$ 0.9 & 
        \underline{140.06} $\pm$ \underline{1.5} &
        \underline{4293.09} $\pm$ \underline{66.44} \\
        
        FAFormer-MHNN & 
        \underline{0.3} $\pm$ \underline{0.003} & 
        4.85 $\pm$ 0.04 & 
        \underline{26.47} $\pm$ \underline{0.2} & 
        \textbf{51.9} $\pm$ \textbf{0.4} & 
        \textbf{73.3} $\pm$ \textbf{0.7} & 
        \textbf{2602.89} $\pm$ \textbf{25.04}\\
        
        Equiformer-MHNN & 
        0.34 $\pm$ 0.003 & 
        \underline{2.48} $\pm$ \underline{0.02} & 
        \textbf{25.57} $\pm$ \textbf{0.2} & 
        \underline{67.19} $\pm$ \underline{0.5} & 
        230.77 $\pm$ 1.8 & 
        102815 $\pm$ 445.433 \\
        \midrule[0.3pt]\bottomrule[1pt]
    \end{tabular}
    }
\end{table}

\cref{tab:qm9_res} presents the Mean Absolute Error (MAE) for six molecular properties from the QM9 dataset. The baseline MHNN, which captures higher-order interactions through hypergraph representations, underperforms compared to standard 2D graph-based models. For example, on $\epsilon_{\textrm{HOMO}}$, MHNN yields an MAE of 55.38 meV, while GIN and GAT achieve 47.67 meV and 51.56 meV, respectively. GIN also records the lowest error on $\mu$ (0.2 meV). These results suggest that modeling higher-order relations alone does not improve the performance of small molecules.

In contrast, incorporating geometric inductive biases leads to significant gains. Both EGNN-MHNN and FAFormer-MHNN consistently reduce MAE across tasks. FAFormer-MHNN achieves the best results in three of the six properties, including $\epsilon_{\textrm{LUMO}}$ (51.9 meV), $\Delta \epsilon$ (73.3 meV), and $\langle R^2 \rangle$ (2602.89 ${a_0}^2$), halving the error compared to the baselines of the MHNN and 2D graph. EGNN-MHNN obtains the lowest MAE in $\alpha$ (2.02 ${a_0}^3$) and competitive results on $\langle R^2 \rangle$ (4293.09 ${a_0}^2$). In particular, Equiformer-MHNN achieves the best performance in $\epsilon_{\textrm{HOMO}}$ (25.57 meV). These findings highlight that while higher-order modeling alone is insufficient, combining topological and geometric priors yields a more powerful and accurate framework for molecular property prediction in the QM9 setting.

\subsubsection{OPV dataset}

\begin{table}[h]
    \centering
    \caption{MAE on the OPV test set.}
    \label{tab:opv_res}
    \renewcommand{\arraystretch}{1.2}
    \resizebox{\textwidth}{!}{
    \begin{tabular}{lcccccccccc}
        \toprule[1pt]\midrule[0.3pt]
        & \multicolumn{4}{c}{Molecular} & \multicolumn{4}{c}{Polymer} \\
        \cmidrule(lr){2-5} \cmidrule(lr){6-9}
        Methods & $\Delta \varepsilon$ & $\varepsilon_{\text{HOMO}}$ & $\varepsilon_{\text{LUMO}}$ & $I_{overlap}$ & $\Delta \varepsilon$ & $\varepsilon_{\text{HOMO}}$ & $\varepsilon_{\text{LUMO}}$ & $O_{\text{LUMO}}$ \\

        Units ($ \downarrow $) & meV & meV & meV & W/mol & meV & meV & meV & meV \\
        \midrule
        GIN & 
        50.45 $\pm$ 0.9 & 
        39.16 $\pm$ 0.5 & 
        53.29 $\pm$ 0.8 & 
        206.53 $\pm$ 3.6 & 
        53.69 $\pm$ 1.0 & 
        61.65 $\pm$ 0.9 & 
        78.48 $\pm$ 1.5 & 
        64.64 $\pm$ 0.6 \\

        GAT & 
        55.8 $\pm$ 0.9 & 
        32.2 $\pm$ 0.5 & 
        46.68 $\pm$ 0.7 & 
        204.03 $\pm$ 4.2 & 
        47.91 $\pm$ 0.83 & 
        58.47 $\pm$ 0.92 & 
        71.84 $\pm$ 1.3 & 
        56.61 $\pm$ 0.7 \\
        
        MHNN & 
        34.02 $\pm$ 0.4 & 
        26.21 $\pm$ 0.4 & 
        24.46 $\pm$ 0.3 & 
        139.58 $\pm$ 2.3 & 
        48.95 $\pm$ 1.1 & 
        \textbf{49.93} $\pm$ \textbf{0.8} & 
        \underline{60.71} $\pm$ \underline{1.1} & 
        \underline{48.41} $\pm$ \underline{0.7} \\
        
        EGNN-MHNN & 
        \underline{28.27} $\pm$ \underline{0.3} & 
        20.97 $\pm$ 0.2 & 
        \underline{20.03} $\pm$ \underline{0.3} & 
        \textbf{99.7} $\pm$ \textbf{1.5} & 
        \underline{45.63} $\pm$ \underline{0.9} & 
        66.67 $\pm$ 1.1 & 
        69.32 $\pm$ 1.2 & 
        67.28 $\pm$ 0.9 \\
        
        FAFormer-MHNN & 
        36.4 $\pm$ 0.6 & 
        \underline{20.5} $\pm$ \underline{0.2} & 
        \textbf{18.84} $\pm$ \textbf{0.3} & 
        \underline{100.52} $\pm$ \underline{1.5} & 
        46.12 $\pm$ 0.9 & 
        54.85 $\pm$ 1.0 & 
        72.05 $\pm$ 1.4 & 
        52.74 $\pm$ 0.8 \\
        
        Equiformer-MHNN & 
        \textbf{28.12} $\pm$ \textbf{0.4} & 
        \textbf{20.24} $\pm$ \textbf{0.2} & 
        20.59 $\pm$ 0.3 & 
        107.346 $\pm$ 1.7 & 
        \textbf{45.42} $\pm$ \textbf{0.9} & 
        \underline{50.08} $\pm$ \underline{0.8} & 
        \textbf{58.17} $\pm$ \textbf{1.0} & 
        \textbf{43.6} $\pm$ \textbf{0.6} \\
        
        \midrule[0.3pt]\bottomrule[1pt]
    \end{tabular}
    }
\end{table}

The OPV dataset includes both small (molecular) and large (polymer) compounds, offering a robust benchmark for evaluating the scalability and generalization of various graph-based representations. \cref{tab:opv_res} highlights the performance (in MAE) of several models in four tasks for each category.

For small molecules, the proposed Equiformer-MHNN outperforms all baselines in two out of four tasks. It achieves the lowest MAE for $\Delta \varepsilon$ (28.12 meV) and $\varepsilon_{\text{HOMO}}$ (20.24 meV). FaFormer-MHNN performs best in $\varepsilon_{\text{LUMO}}$ with 18.84 meV, while EGNN-MHNN leads in $I_{\text{overlap}}$. When applied to molecular tasks, incorporating symmetry awareness into the hypergraph consistently outperforms 2D graphs and models that rely solely on higher-order interactions.

In contrast, for larger polymer molecules, Equiformer-MHNN continues to demonstrate strong performance, outperforming all other models in three of four tasks. It achieves the best MAE for $\Delta \varepsilon$ (45.42 meV), $\varepsilon_{\text{LUMO}}$ (58.17 meV) and $O_{\text{LUMO}}$ (43.6 meV), with a close second for $\varepsilon_{\text{HOMO}}$ (50.08 meV). Although EGNN-MHNN and FAFormer-MHNN do not surpass Equiformer-MHNN, they still significantly outperform traditional 2D GNNs across all polymer-related tasks, highlighting the importance of incorporating geometric and equivariant representations when modeling complex macromolecules.

The baseline MHNN, which models high-order interactions without geometric inductive bias, achieves a moderate performance boost over GIN and GAT for small molecules (e.g., 24.46 meV in $\varepsilon_{\text{LUMO}}$ vs 53.29 and 46.68 meV for GIN and GAT, respectively). However, its improvements diminish in the polymer regime, where long-range dependencies and complex geometry require more expressive representations.

\subsubsection{PCQM4Mv2}

\begin{table}[h]
    \centering
    \caption{MAE on the PCMQM4Mv2 test set in meV.}
    \renewcommand{\arraystretch}{1.2}
    \label{tab:pcqm_res}
    \begin{tabular}{lc}
        \toprule[1pt]\midrule[0.3pt]
        Model & gap ($ \downarrow $) \\
        \midrule
        GIN & 117.65 $\pm$ 0.23 \\

        GAT & 116.93 $\pm$ 0.21 \\

        MHNN & \underline{108.11} $\pm$ \underline{0.25} \\
        
        EGNN-MHNN & \textbf{98.45} $\pm$ \textbf{0.2} \\
        
        \midrule[0.3pt]\bottomrule[1pt]
    \end{tabular}
\end{table}

 \cref{tab:pcqm_res} reports the MAE for the prediction of the HOMO-LUMO gap in the PCQM4Mv2 test set. The baseline models, GIN and GAT, achieve errors of 117.65 meV and 116.93 meV, respectively. MHNN reduces the error to 108.11 meV by incorporating high-order molecular interactions through hypergraph representations. EGNN-MHNN further improves the performance to 98.45 meV by integrating 3D geometric features with equivariant message passing. These results indicate that geometric inductive biases and spatial information are beneficial for learning quantum chemical properties in large-scale molecular graphs.

\subsubsection{Molecule3D}

\begin{table}[h]
    \centering
    \caption{MAE on the Molecule3D test set in meV.}
    \renewcommand{\arraystretch}{1.2}
    \label{tab:mole_res}
    \begin{tabular}{lc}
        \toprule[1pt]\midrule[0.3pt]
        Model & gap ($ \downarrow $) \\
        \midrule
        GIN & 129.61 $\pm$ 0.25 \\

        GAT & 137.22 $\pm$ 0.36 \\

        MHNN & \textbf{117.55} $\pm$ \textbf{0.28} \\
        
        EGNN-MHNN & \underline{122.25} $\pm$ \underline{0.24} \\
        
        \midrule[0.3pt]\bottomrule[1pt]
    \end{tabular}
\end{table}

\cref{tab:mole_res} shows the MAE results in the Molecule3D test set. GIN and GAT yield errors of 129.61 meV and 137.22 meV, respectively. MHNN achieves the lowest error of 117.55 meV, demonstrating the effectiveness of high-order interaction modeling to capture structural dependencies. EGNN-MHNN obtains a slightly higher error of 122.25 meV, suggesting that the added geometric modeling does not consistently improve performance in datasets with high conformational flexibility.

\section{Conclusion}

In this work, we have explored the integration of symmetry-aware features into hypergraph representations for molecular modeling. Our approach focuses on a simple yet effective strategy of preparing node features through an embedding that combines both symmetry-aware geometric representations and invariant scalar information. We experimented with several equivariant techniques, including spatial domain modeling, frame averaging, and Fourier domain methods. Although we also attempted to directly modify the hypergraph message passing framework, this approach proved to be not only challenging but also yielded poor results, highlighting the elegance and practicality of our embedding-based strategy.

The results demonstrate that hypergraphs consistently outperform pairwise graph approaches on larger molecules, showcasing their scalability. Additionally, incorporating symmetry-aware features significantly improves model performance, emphasizing the importance of capturing both high-order interactions and geometric consistency. Overall, our model demonstrates strong generalization capabilities to large graphs while maintaining a simple and robust framework.
\section{Limitations and Future work}

In molecular systems, various higher-order interactions beyond conjugated bonds are important, such as those explored by ETNNs~\cite{battiloro2024n}. Furthermore, SE3Set~\cite{wu2024se3set} employs a fragmentation method using a BFS-like algorithm to identify connected subgraphs, which could be valuable for capturing complex interactions. Our experiments with using rings as higher-order interactions resulted in significantly worse performance compared to conjugated bonds, highlighting the need for further exploration of effective high-order interactions.

Moreover, while several state-of-the-art equivariant models outperform Equiformer on certain tasks with better computational and parameter efficiency, our work demonstrates a simple and flexible framework where node features for hypergraph neural networks are initialized with embeddings from equivariant models. This design offers a straightforward plug-and-play capability, allowing integration with other more advanced equivariant architectures to further enhance performance and scalability. Future work will explore the incorporation of these models to unlock better efficiency and generalization.

\bibliography{main}
\bibliographystyle{unsrt}

\newpage
\appendix

\section{Message Passing}

Message Passing Neural Networks (MPNNs) play a fundamental role in learning node representations by propagating information along graph edges. As inherently permutation-invariant architectures, MPNNs effectively capture relational structures in graph-structured data, making them particularly well-suited for applications such as molecular modeling ~\cite{gilmer2017neural}.

Given a graph \(\mathcal{G} = (\mathcal{V}, \mathcal{E})\) with nodes \(v_i \in \mathcal{V}\) and edges \(e_{ij} \in \mathcal{E}\), each node \(v_i\) is associated with a feature vector \(\mathbf{h}_i \in \mathbb{R}^{c_n}\) and each edge \(e_{ij}\) with a feature vector \(\mathbf{a}_{ij} \in \mathbb{R}^{c_e}\), where \(c_n\) and \(c_e\) represent the dimensionalities of the node and edge features, respectively. The nodes representation are iteratively updated by:
\begin{equation}
\begin{aligned}
    \mathbf{m}_{ij} &= \phi_{e}(\mathbf{h}^l_i, \mathbf{h}^l_j, \mathbf{a}_{ij}) \\ 
    \mathbf{m}_i &= \text{AGGREGATE} \bigg( \{{\mathbf{m}_{ij}\}}_{j\in\mathcal{N}(i)} \bigg) \\
    \mathbf{h}^{l+1}_i &= \phi_{u}(\mathbf{h}^l_i, \mathbf{m}_i)
\end{aligned}
\end{equation}
where \(\mathcal{N}(i)\) denotes the set of neighbors of node \(v_i\), and the AGGREGATE function is a permutation-invariant operation over the neighbors (e.g., summation). The functions \(\phi_m\) and \(\phi_e\) are the message computation and the feature update function, respectively, typically parameterized by multilayer perceptrons (MLP). To obtain the final graph representation, a permutation-invariant aggregator is applied to the final hidden states of all nodes. However, this operation itself does not inherently preserve the E(3) equivariance.

\section{Equivariance}

Equivariance is a property of functions that ensures that information is preserved under transformations from a group \( G \). A function \( \phi: \mathcal{X} \to \mathcal{Y} \) is equivariant with respect to \( G \) if for all \( g \in G \) and \( x \in \mathcal{X} \), the following holds:
\[
\phi(g \cdot x) = g \cdot \phi(x).
\]
This property ensures that the output transforms in the same way as the input under transformations like rotation, translation, or scaling. In molecular modeling, the equivariance with these symmetries (e.g., the rotation group \( SO(3) \)) is crucial because molecules retain their physical properties regardless of their orientation or position. 



\section{Hypergraph}

A \textit{hypergraph} is a generalization of a graph where an edge, called a \textit{hyperedge}, can connect any number of nodes. Formally, a hypergraph is defined as: 
\[
\mathcal{H} = (\mathcal{V}, \mathcal{E}),
\] 
where \(\mathcal{V}\) is the set of nodes and \(\mathcal{E} \subseteq 2^{\mathcal{V}} \setminus \{\emptyset\}\) is the set of hyperedges, each representing a subset of nodes. Unlike traditional graphs that capture pairwise relationships, hypergraphs model higher-order interactions, making them ideal for complex systems such as protein interactions
~\cite{murgas2022hypergraph} and chemical reactions~\cite{schwaller2020predicting}.  

Hypergraph-based models extend GNNs to process hyperedges, allowing better representation of structured data. In molecular chemistry, hypergraphs effectively model \textit{conjugated structures}, where delocalized electrons form multi-atomic interactions crucial for optoelectronic properties. Recent studies demonstrate that HGNNs outperform traditional GNNs and 3D-based models in predicting molecular properties, offering a powerful approach for data-scarce applications like organic semiconductor design.

\begin{table}
    \centering
    \caption{Model Architecture Overview}
    \renewcommand{\arraystretch}{1.2}
    \label{tab:model_overview}
    \resizebox{\textwidth}{!}
    {
        \begin{tabular}{lcccc}
            \toprule[1pt]\midrule[0.3pt]
            Model & No.parameters & No.layers & No. attention heads & Hidden dimension\\
            \midrule
            GIN & 965K & 2 & - & 256 \\
            GAT & 3.7M & 2 & 4 & 256 \\
            MHNN & 2.5M & 2 & - & 256 \\
            EGNN & 844K & 2 & 8 & 256 \\
            FAFormer & 3.2M & 2 & 2 & 256 \\
            Equiformer & 13M & 1 & 1 & 256 \\
            \midrule[0.3pt]\bottomrule[1pt]
        \end{tabular}
    }
\end{table}

\section{Model details}
\label{appendix:model_overview}

\cref{tab:model_overview} outlines the architectural configurations of the models used in our experiments, including the number of parameters, layers, attention heads, and hidden dimensions.

\end{document}